\documentclass[10pt,twocolumn,letterpaper]{article}

\usepackage[accsupp]{axessibility}  
\usepackage[pagenumbers]{wacv} 

\usepackage{amsmath}
\usepackage{amssymb}
\usepackage{mathtools}
\usepackage{amsthm}

\usepackage{pgfplots}
\pgfplotsset{compat=1.18}
\usepackage{verbatim}
\usepackage{graphicx}
\usepackage{pifont}
\usepackage{colortbl} 
\usepackage[table,dvipsnames,x11names]{xcolor} 
\usepackage{xspace}
\usepackage{multirow} 
\usepackage{stfloats} 
\usepackage{pifont}

\usepackage{makecell}

\theoremstyle{plain}

\theoremstyle{definition}

\theoremstyle{remark}

\definecolor{gray}{gray}{0.9}
\definecolor{aquamarine}{rgb}{0.5, 1.0, 0.83}
\definecolor{aqua}{rgb}{0.0, 1.0, 1.0}
\definecolor{pink}{rgb}{1.0, 0.75, 0.8}
\definecolor{lightbrightturquoise}{rgb}{0.68, 0.85, 0.9}
\definecolor{brightturquoise}{rgb}{0.03, 0.91, 0.87}

\newcommand{\dataset}{MarineLife-16K\xspace}

\definecolor{wacvblue}{rgb}{0.21,0.49,0.74}
\usepackage[pagebackref,breaklinks,colorlinks,allcolors=wacvblue]{hyperref}

\def\wacvPaperID{384} 
\def\confName{WACV}
\def\confYear{2027}

\title{\dataset: An Attribute-Rich Video-Text Dataset for Motion-Aware Marine Video Understanding}

\author{
Quang-Trung Truong\textsuperscript{1,2}\quad 
Thanh-Nghia Nguyen\textsuperscript{3}\quad
Phu-Phong Vo\textsuperscript{3}\quad 
Jaeyeon Kim\textsuperscript{2}\quad
Cheng Chen\textsuperscript{2} \\
Jie Zhang\textsuperscript{2}\quad 
Binh-Son Hua\textsuperscript{4}\quad 
Duc Thanh Nguyen\textsuperscript{5}\quad 
Ivor Tsang\textsuperscript{2, 6}\quad 
Sai-Kit Yeung\textsuperscript{1} \\[1ex]
\textsuperscript{1}Hong Kong University of Science and Technology, Hong Kong SAR \\
\textsuperscript{2}Institute of Advanced Intelligence and Computing (IAIC),\\
Agency for Science, Technology and Research (A*STAR), Singapore, Singapore\quad \\ 
\textsuperscript{3}University of Information Technology, Vietnam National University, Ho Chi Minh city, Vietnam\\
\textsuperscript{4}Trinity College Dublin, Ireland \quad
\textsuperscript{5}Deakin University, Australia\\
\quad
\textsuperscript{6}Nanyang Technological University, Singapore, Singapore\\
\href{https://marinelife-16k.hkustvgd.com}{\texttt{marinelife-16k.hkustvgd.com}}
}

\begin{document}
\maketitle

\begin{figure*}[b]
    \centering
        \includegraphics[width=0.9\linewidth]{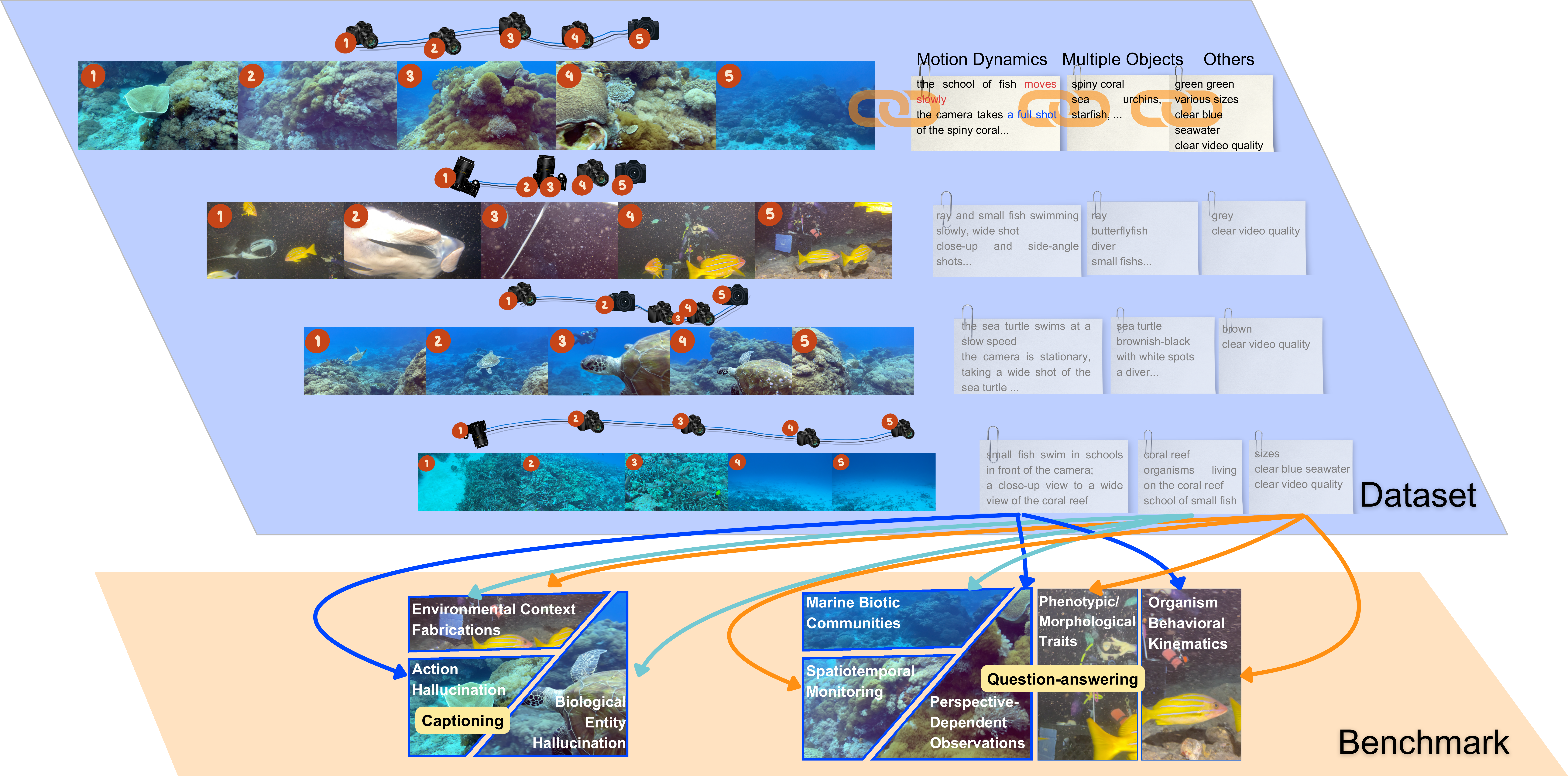}
        \caption{\textbf{Annotation Highlights.} We build a dataset of 2,000 video–text pairs, containing fine-grained motion dynamics, including diverse object and camera poses, captured across 7 countries. \textbf{Top}: We show video–attribute pairs, highlighting \textcolor{red}{object motion} and \textcolor{blue}{camera motion} to reveal the motion dynamics in marine scenarios. \textbf{Bottom}: We benchmark our dataset on video understanding tasks, ranging from captioning to question-answering. We show how \textcolor{blue}{Motion Dynamics}, \textcolor{cyan}{Multiple Objects} and \textcolor{orange}{Others} contribute to different dimensions.}
        \label{fig:teaser}
\end{figure*}

\begin{abstract}
Marine video understanding is crucial for a wide range of real-world applications. However, existing marine video datasets suffer from limitations, including the limited number of data samples, fine-grained text descriptions, and detailed visual attributes. Visual attributes, capturing appearance (color, texture) and geometry (shape, size), comprise a significant portion of scene information, yet they are often confounded by camera motion. To address these limitations, we introduce a large-scale, video-text dataset featuring rich, in-the-wild visual attributes tailored for complex marine environments. Our dataset comprises \textbf{2,000} video-text pairs annotated with \textbf{13} attributes covering up to four objects per scene, object visual/geometric properties, behaviors, environmental context, and camera motion, spanning diverse lighting conditions, times of day, and 7 countries. From these annotations, we generate \textbf{16,080} video-question pairs (\textbf{12,080} multiple-choice, \textbf{4,000} open-ended). Furthermore, we benchmark both proprietary and open-source video-LLM models on video question answering and video captioning tasks. We find that (i) proprietary models often achieves superior performance; (ii) VLMs handle object motion well but struggle with complex camera motion; (iii) closed-source models struggle to perceive camera motion; and (iv) VLMs struggle to reason in static cameras.
\end{abstract}

\begin{figure*}[t]
    \centering
        \includegraphics[width=0.8\linewidth]{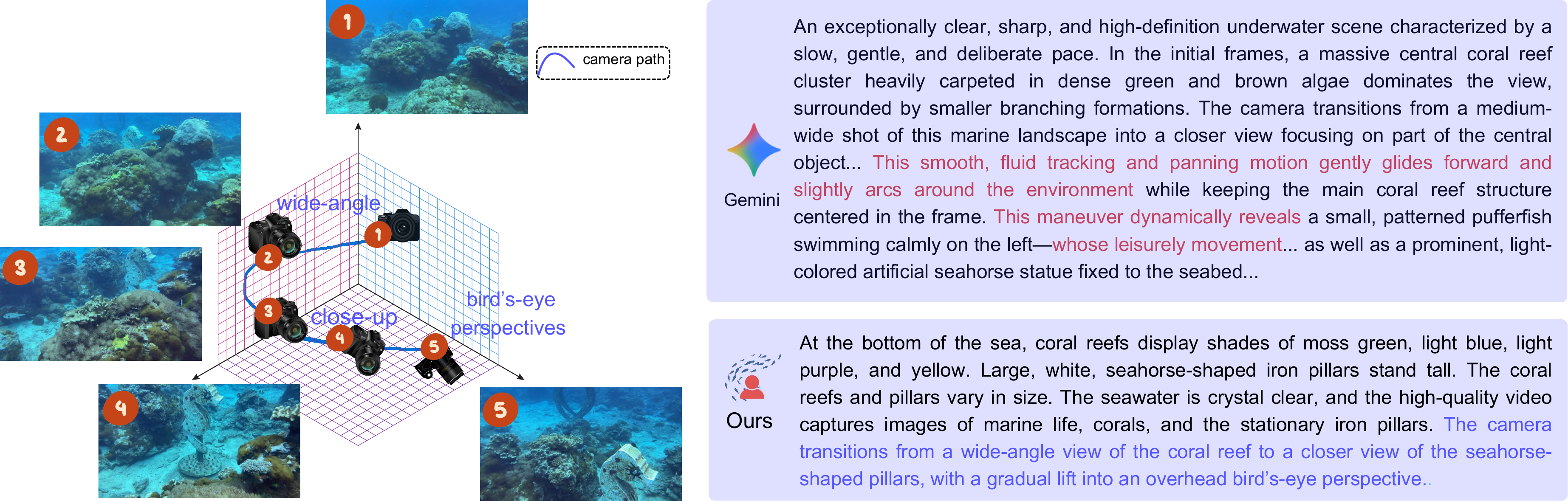}
        \caption{Illustration of motion hallucination. The \textcolor{red}{red} texts represent hallucinated camera motion by Gemini 3.1~\cite{gemini31docs}, demonstrating failure to capture camera movement. We capture camera motion ground truths with {\color{Red}\ding{182}}-{\color{Red}\ding{183}}:\textcolor{blue}{wide-angle}, {\color{Red}\ding{184}}-{\color{Red}\ding{185}}:\textcolor{blue}{close-up}, {\color{Red}\ding{186}}:\textcolor{blue}{bird's-eye perspective}.}
        \label{fig:motion_halluc_gemini}
\end{figure*}

\section{Introduction}
\label{sec:intro}
Marine video analysis presents unique challenges to computer vision, particularly marine life analysis, e.g., fish, whose complex behaviors are intertwined with functional purposes~\cite{pitcher2012behaviour, moyle2000fishes}. For instance, schooling behavior serves critical ecological functions, including optimized orientation, hunting, predator confusion, and risk mitigation. Conversely, when individual fish detach from a school to navigate around reefs, their actions often signal a shift toward localized foraging, shelter-seeking, or resting.
Capturing these behaviors requires changing from object-centric tracking to motion-centric video understanding~\cite{shuai2025free, lin2026towards}.

Current video-text benchmarks remain limited when applied to these real-world ecological contexts. Existing datasets focus on global video summarization~\cite{zhou2024streaming}---describing actions at a broad level without localizing specific timestamps. In contrast, video Question Answering (video-QA) has emerged as a powerful paradigm to evaluate fine-grained reasoning across different temporal windows~\cite{yuanvideorefer}. However, existing video-QA benchmarks suffer from a lack of motion diversity and fail to challenge models on complex, real-world temporal dynamics.

For QA, it requires datasets that are richly and carefully annotated across different aspects like fact-centric QAs~\cite{shaar2026movierecapsqa}, relationship and motion questions~\cite{dai2025towards}, focusing on fine-grained attributes. Furthermore, LLM captioning models are often redesigned for QAs, but struggle to perform grounding. To address those limitations, we introduce a novel video-text benchmark dataset for marine video analysis, evaluating dense video captioning and complex motion-aware QA. 

Although video-language models (VLMs) are powerful at generating text, they still suffer from hallucinations ~\cref{fig:motion_halluc_gemini}. Leveraging VLMs for captioning complex videos such as marine footage remains ineffective. In particular, MSC\textcolor{BrickRed}{~\cite{truong2025msc}} introduces the two-stage video captioning pipeline, aiming to reduce object hallucinations, but fails to detect motion hallucinations in videos. To solve this, we propose a human-in-the-loop captioning engine that reduces motion hallucinations through manual, fine-grained attribute annotations. Compared with MSC~\cite{truong2025msc}, our dataset is 5× larger in scale and extracts motion-aware attributes. We compare our \dataset with existing video-text paired datasets in ~\cref{table:description_datasets}.

In summary, we make the following contributions in our work.
\begin{itemize}
    \item We propose a scalable caption engine to reduce hallucinations, using a human-in-the-loop workflow annotated by domain experts.
    \item We introduce a large-scale motion-driven video dataset with 2,000 video–text pairs, featuring 13 fine-grained attributes for the marine domain.
    \item We develop a benchmark for question answering (QA) tasks, with 16K question–answer pairs, to evaluate eight open-source models and four closed-source models. We observe that VLM models struggle with camera-motion understanding.
\end{itemize}

\section{Related Works}
\label{sec:related_works}
\paragraph{Marine Video Understanding using LLMs.}
Marine video analysis has recently attracted considerable interest from the computer vision community. Recent works such as MVK~\cite{truong2023marine} and MSC~\cite{truong2025msc} leverage LLMs to understand marine scenarios. 
More broadly, models like OceanGPT~\cite{bi2024oceangpt}, VideoRefer Suite~\cite{yuanvideorefer}, and Video-LLaMA3~\cite{zhang2025videollama} integrate spatio-temporal video features and language within a unified embedding space. Specifically, these frameworks map sequential visual tokens into an LLM text space to generate open-ended responses. However, existing LLM architectures still struggle with fine-grained temporal comprehension, which can cause hallucinations of complex camera motion and object motion dynamics in challenging underwater scenarios.

\paragraph{Video Question Answering.}
There has been a growing interest in multimodal VideoQA benchmarks like EgoSchema \cite{mangalam2023egoschema}, where models must tightly integrate long-form visual and textual contexts to reason across modalities. Moving beyond generic video understanding, recent benchmarks introduce specialized dimensions of temporal and structural complexity. For instance, HD-EPIC \cite{perrett2025hd} establishes fine-grained, multi-modal egocentric alignments to evaluate action and gaze anticipation over extended timelines, while MovieRecapsQA \cite{shaar2026movierecapsqa} shifts the focus toward long-form narrative comprehension by grounding open-ended QA pairs in atomic movie facts.

\begin{table*}
    \centering
    \scalebox{0.7}{
        \begin{tabular}{p{3.8cm} p{0.8cm} p{1.7cm} p{1.2cm} p{1cm} p{1.7cm} p{2cm} p{1cm} p{1cm} p{1.1cm} p{1.7cm}}
            \toprule
            Dataset & Time & Domain & \#Sentence & Caption & \#Video-text & Cap Source & \multicolumn{3}{c}{Attributes} & Prompt \\
            \cmidrule(lr){8-10}
            & (hours) & & & Length & Pairs & & Object & Motion & Camera & Complexity \\
            \midrule
            WebVid~\cite{bain2021frozen} & 13K & open & - & 12 & 2.5M & generated & \ding{55} & \ding{55} & \ding{55} & low \\
            InternVid-VTT~\cite{wang2024internvid} & 76K & open & 1 & 10-20 & 7.1M & generated & \ding{55} & \ding{55} & \ding{55} & medium \\
            MiraData~\cite{ju2024miradata} & 16K & - & - & - & - & generated & \ding{55} & \ding{55} & \ding{55} & low \\
            MovieBench~\cite{wu2025moviebench} & 69.2 & movie & - & - & - & generated & \ding{51} & \ding{51} & \ding{51} & medium \\
            MSR-VTT~\cite{xu2016msr} & 41.2 & open & 1 & 9.3 & 200K & manual & \ding{55} & \ding{55} & \ding{55} & low \\
            EPIC-KITCHENS-100~\cite{damen2022rescaling} & 100 & cooking & 1 & 3 & 19.8K & manual & \ding{55} & \ding{55} & \ding{55} & low \\
            InternVid~\cite{wang2024internvid} & 760K & open & 1 & 17.6 & 234M & generated & \ding{55} & \ding{51} & \ding{55} & medium \\
            \midrule
            MovieRecapsQA~\cite{shaar2026movierecapsqa} & 107 & movie & - & - & 60 & generated & \ding{51} & \ding{51} & \ding{55} & high \\
            MOMA-QA~\cite{dai2025towards} & 147 & open & - & - & 1412 & manual & \ding{51} & \ding{51} & \ding{55} & high \\
            \midrule
            MVK~\cite{truong2023marine} & 12.3 & underwater & 1 & 9.8 & 1378 & generated & \ding{55} & \ding{55} & \ding{55} & low \\
            MSC~\cite{truong2025msc} & 3.6 & underwater & 2.27 & 27.4 & 396 & manual & \ding{51} & \ding{51} & \ding{55} & high \\
            \rowcolor{lightgray} \textsc{\dataset} (Ours) & 18.48 & underwater & 3.4 & 44 & 2000 & manual & \ding{51} & \ding{51} & \ding{51} & high \\
            \bottomrule
        \end{tabular}
    }
    \caption{Comparison with existing mainstream video caption datasets. \dataset features extensive manual annotations for underwater video data, incorporating several attributes, i.e., object awareness, object motion, and camera motion in a hierarchical manner. Note, MovieRecapsQA~\cite{shaar2026movierecapsqa} and MOMA-QA~\cite{dai2025towards} datasets are designed for QA tasks, but our dataset is larger than the QA counterparts.}
    \label{table:description_datasets}
    \vspace{-0.2in}
\end{table*}

\begin{figure}[t]
    \centering
    \includegraphics[width=0.9\linewidth]{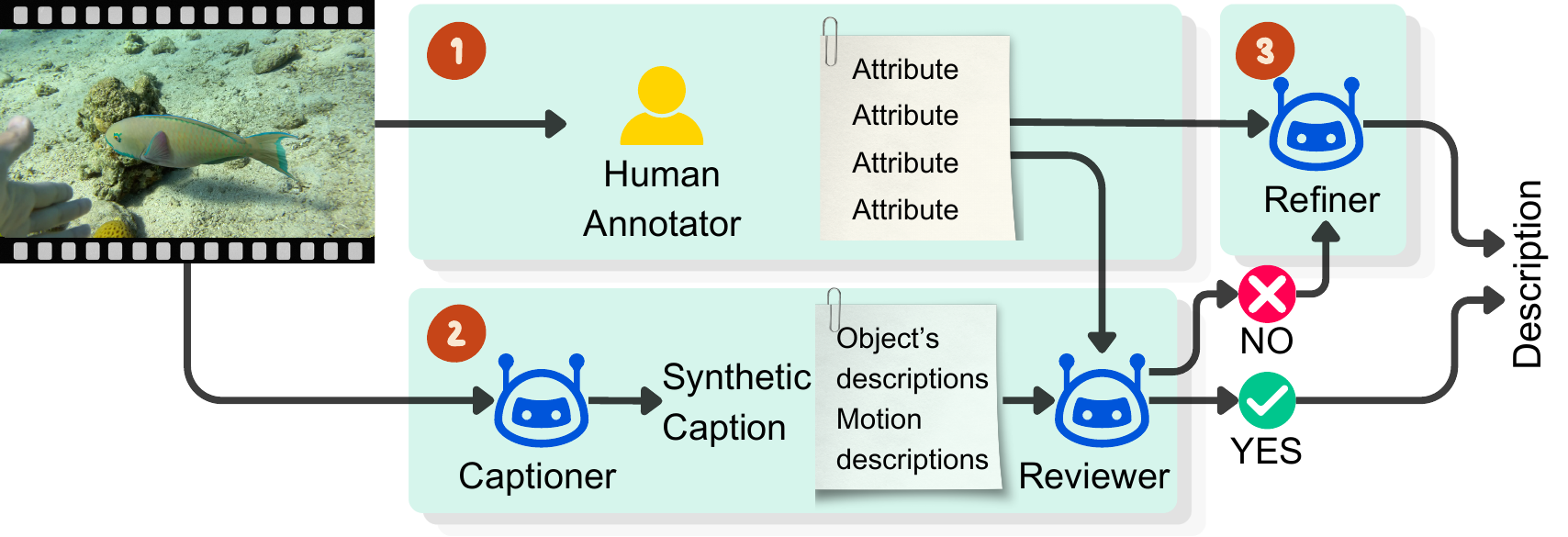}
    \caption{A caption engine. We introduce a three-stage, human-in-the-loop caption engine that utilizes a multi-agent framework to generate high-quality video captions. {\color{Red}\ding{182}} Fine-grained attributes are manually annotated by human experts. {\color{Red}\ding{183}} A LMM captioner agent inputs video sequences to generate initial synthetic captions, which a reviewer agent subsequently scores against the human-verified attributes {\color{Red}\ding{184}} A refiner agent filters the outputs, correcting low-confidence synthetic captions to ensure factual alignment.}
    \label{fig:data_engine}
    \vspace{-0.2in}
\end{figure}

\paragraph{Video-Text Datasets.}
Recently, many large-scale video–text datasets, e.g., Koala-36M~\cite{wang2025koala}, HOIGen-1M~\cite{liu2025hoigen}, and MiraData~\cite{ju2024miradata} have leveraged automatic captioning systems (e.g., GPT-4V) to generate video captions. In addition, datasets such as MVK~\cite{truong2023marine} and MSC~\cite{truong2025msc} have recently been introduced for the marine domain, providing insights into VLMs’ marine understanding capabilities, including video retrieval and video grounding.

\section{\dataset - A real-world video-text dataset with fine-grained annotations}
\label{sec:UTV}

\subsection{Data engine.}
We present a scalable, three-stage marine caption engine that leverages human-in-the-loop attributes and multi-agent alignment scoring to generate high-fidelity, factually grounded video descriptions, illustrated in~\cref{fig:data_engine}. We use LLM, i.e., GPT-5.5~\cite{gpt} for dataset construction steps.

\begin{itemize}
    \item \textit{Step 1: Manual attributes.} Three domain experts extract discrete marine attributes from raw videos.
    \item \textit{Step 2: Multi-agent-based synthetic caption.} We leverage LLM agents to caption marine videos. Specifically, an LMM Captioner is used to draft text descriptions; an LMM Reviewer scores the semantic alignment between the manual attributes and the synthetic caption, yielding a matching score ($S_c$), where $S_c \in [0, 5]$ .
    \item \textit{Step3: Final caption.} If $S_c < \tau$, a refiner incorporates human attributes into a synthetic caption to output the final caption; otherwise, it accepts the synthetic caption. We set $\tau$ to 3.5 in our work.
\end{itemize}

\begin{figure*}[t]
    \centering
        \includegraphics[width=0.85\linewidth]{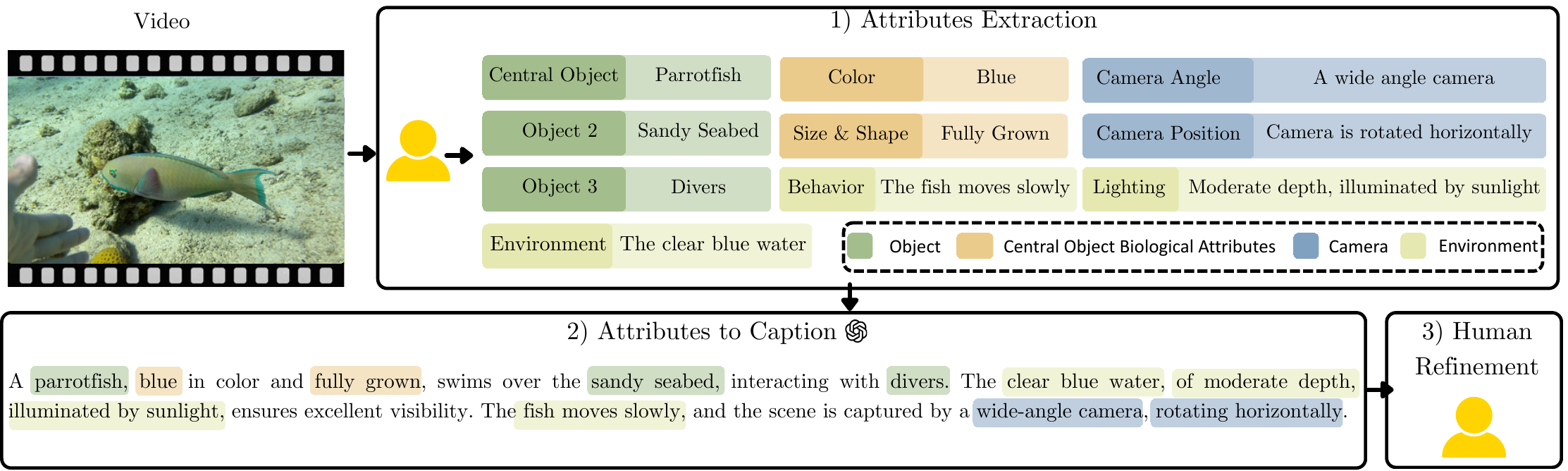}
        \caption{An example of annotated attributes.}
        \label{fig:datapipeline}
        \vspace{-0.2in}
\end{figure*}

\subsection{Collection and annotation}
\paragraph{Overview.}
Here, we introduce a real-world dataset, including 2,000 video-text pairs in the marine domain. The example of our \dataset is illustrated in ~\cref{fig:datapipeline}. 
\paragraph{Fine-grained attributes.}
We also provide fine-grained annotations, including \texttt{central object} and up to 3 \texttt{additional objects}, \texttt{environment}, \texttt{lighting}, \texttt{video clarity}, \texttt{motion/behavior}, and camera attributes, e.g., \texttt{angle} and \texttt{position} for the collected videos. The captioning annotation begins with manually watching each video entirely to identify a central object and any other relevant objects. For each \texttt{central object}, we further annotate its appearance attributes, including \texttt{texture}, \texttt{size}, and \texttt{shape\&color}. We allow up to 4 objects in each video to be extracted and described in the video's caption. It is important to provide insights of the underwater environment for future ecological studies. To do so, we further describe the environmental factors from the videos, including lighting conditions, video clarity, any movement or behavior observed, and the camera attributes (position, movement, and sharpness), which we describe in detail below. This captioning process takes around 10 minutes per video, excluding the time required to watch the content. 

\paragraph{Multi-object.} We bound the maximum object density at four objects per clip to maintain descriptive quality. The targeted taxonomy includes common marine entities, such as mobile fauna (fish, human divers), benthic flora (algae, seaweed, seagrass), and foundational substrate organisms (coral and encrusting sponges). 

\paragraph{Object motion taxonomy and behavioral categorization.} To comprehensively model temporal dynamics, we formalize a fine-grained taxonomy of marine object motions. We decompose behavioral patterns into active kinematics, metabolic functions, and survival strategies. Crucially, our data engine formalizes both the empirical movement speed/behavior and ecological context. We categorize these dynamic visual events into four primary behavioral dimensions:
(1) Locomotion and kinematics: tracks fine-grained motion of fins, tails, and body musculature during swimming. (2) Behaviors: encompass fine-grained feeding actions (hunting, stalking, chewing, swallowing, or filter-feeding via suction, biting, and snapping). (3) Survival strategies: models resting/hiding states inside localized substrates (e.g., rocks, coral reefs, vegetation, caves) alongside active defenses such as morphological camouflage, color-changing dynamics. (4) Socio-ecological dynamics: captures coordinated group dynamics (e.g., collective schooling/shoaling) and reproductive events like live birth.

\paragraph{Camera motion dynamics.} We describe the dynamic camera using three attributes: \textit{camera position}, \textit{camera angle}, and \textit{video sharpness}. For camera position, we specify whether the camera captures the \textit{full scene, a single viewing angle, or only a specific part of the central object} (e.g., a portion of the main object). For video sharpness, we categorize the footage as either clear or blurry. We also characterize the camera angle and movement. The camera may perform a horizontal pan (turning left/right), a vertical tilt (moving up/down), or a dolly/translation movement in which the camera shifts position while keeping the main object as the reference center.

\subsection{Video data quality.}

\paragraph{Increasing data diversity.}
We selected videos, recorded from 27 different locations and at different times, spanning 7 countries. We curated the videos from two datasets, MVK~\cite{truong2023marine} and MSC~\cite{truong2025msc} datasets.

\paragraph{Quality control.} We performed quality control by having six human reviewers assess 2,000 text captions.

\paragraph{Fine-grained attributes with motion dynamics.}
Our dataset contains fine-grained attributes, aiming to reflect the primary challenges of real-world videos, outlined in Table~\ref{tab:video_attributes}.

\subsection{Motion-aware QA Generation}
For multiple-choice questions, we use fine-grained attributes as answers and use an LLM prompt to generate the remaining answer choices. For open-ended QA questions, we use video captions as facts to create question-answer pairs, evaluating the VLM’s ability to understand camera motions. Specifically, the video QA task is designed to evaluate a model’s capability to decode the functional relationship between camera movement and the corresponding object's attributes, including motion, appearance, and geometry. We use GPT-5.5~\cite{gpt} as LLM judge. Finally, we generate 16,080 video-question pairs, including 12080 multiple choice questions and 4,000 open-ended questions for the collection of 2,000 videos.

\begin{table}[t]
\centering
\definecolor{color_past}{HTML}{FFF0F5}     
\definecolor{shape_past}{HTML}{FFFACD}     
\definecolor{light_past}{HTML}{E0FFFF}     
\definecolor{motion_past}{HTML}{E6E6FA}    
\definecolor{cam_past}{HTML}{FF9A86}       
\definecolor{bg_past}{HTML}{E8F8F5}        
\definecolor{behaviour}{HTML}{FFD400}
\definecolor{obj}{HTML}{FFE4E1}
\definecolor{obj_mo}{HTML}{9AD872}

\scalebox{0.65}{
\begin{tabular}{p{8.5cm}|p{2.3cm}} 
\toprule
Video Challenge & Visual Attribute \\
\midrule
\multicolumn{2}{l}{\textbf{Object-Centric}} \\
\midrule
\textit{Heterogeneous Object.} Object regions have distinct colors. & 
\colorbox{color_past}{Color} \\
\addlinespace
\textit{Shape Complexity.} The object has complex boundaries, such as thin parts and holes. & 
\colorbox{shape_past}{Size/Shapes} \\
\addlinespace
\textit{Appearance Change.} Noticeable appearance variation, due to illumination changes and relative camera-object rotation. & 
\colorbox{light_past}{Lighting}, \colorbox{cam_past}{Camera Motion}, \colorbox{shape_past}{Size/Shape} \\
\addlinespace
\textit{Low Resolution.} The ratio between the average object bounding-box area and the image area is smaller than 0.1. & 
\colorbox{shape_past}{Size/Shape} \\
\addlinespace
\midrule
\multicolumn{2}{l}{\textbf{Object \& Camera Pose}} \\
\midrule
\textit{Fast-Motion.} The average, per-frame object motion, computed as the Euclidean distance between centroids, is larger than 20 pixels. & 
\colorbox{behaviour}{Behavior} \\
\addlinespace
\textit{Camera-Shake.} Footage displays non-negligible vibrations. & 
\colorbox{cam_past}{Camera Motion} \\
\addlinespace
\textit{Motion Blur.} The object has fuzzy boundaries due to fast motion. & 
\colorbox{light_past}{Lighting}, \colorbox{behaviour}{Behavior} \\
\addlinespace
\textit{Out-of-view.} Object is partially clipped by the image boundaries. & 
\colorbox{obj}{Object}, \colorbox{behaviour}{Behavior}\\
\addlinespace
\textit{Occlusion.} Object becomes partially or fully occluded. & 
\colorbox{obj}{Object} \\
\addlinespace
\textit{Scale-Variation.} The relative area ratio between any two instances of the target object is restricted to less than 0.5. & 
\colorbox{obj_mo}{Object Motion}, \colorbox{behaviour}{Behavior} \\
\addlinespace
\textit{Interacting Objects.} The target object is an ensemble of multiple, spatially-connected objects (e.g., school of fish). & 
\colorbox{behaviour}{Behavior} \\
\midrule
\multicolumn{2}{l}{\textbf{Dynamic Background Context}} \\
\midrule
\textit{Dynamic Background.} Background regions move or deform. & 
\colorbox{bg_past}{Background} \\
\bottomrule
\end{tabular}}
\caption{Taxonomy of video challenges and link to visual attributes. We extend attributes~\cite{PerazziPMGGS16} to video-text understanding.}
\label{tab:video_attributes}
\vspace{-0.2in}
\end{table}

\begin{figure*}
    \centering
    \includegraphics[width=0.8\textwidth]{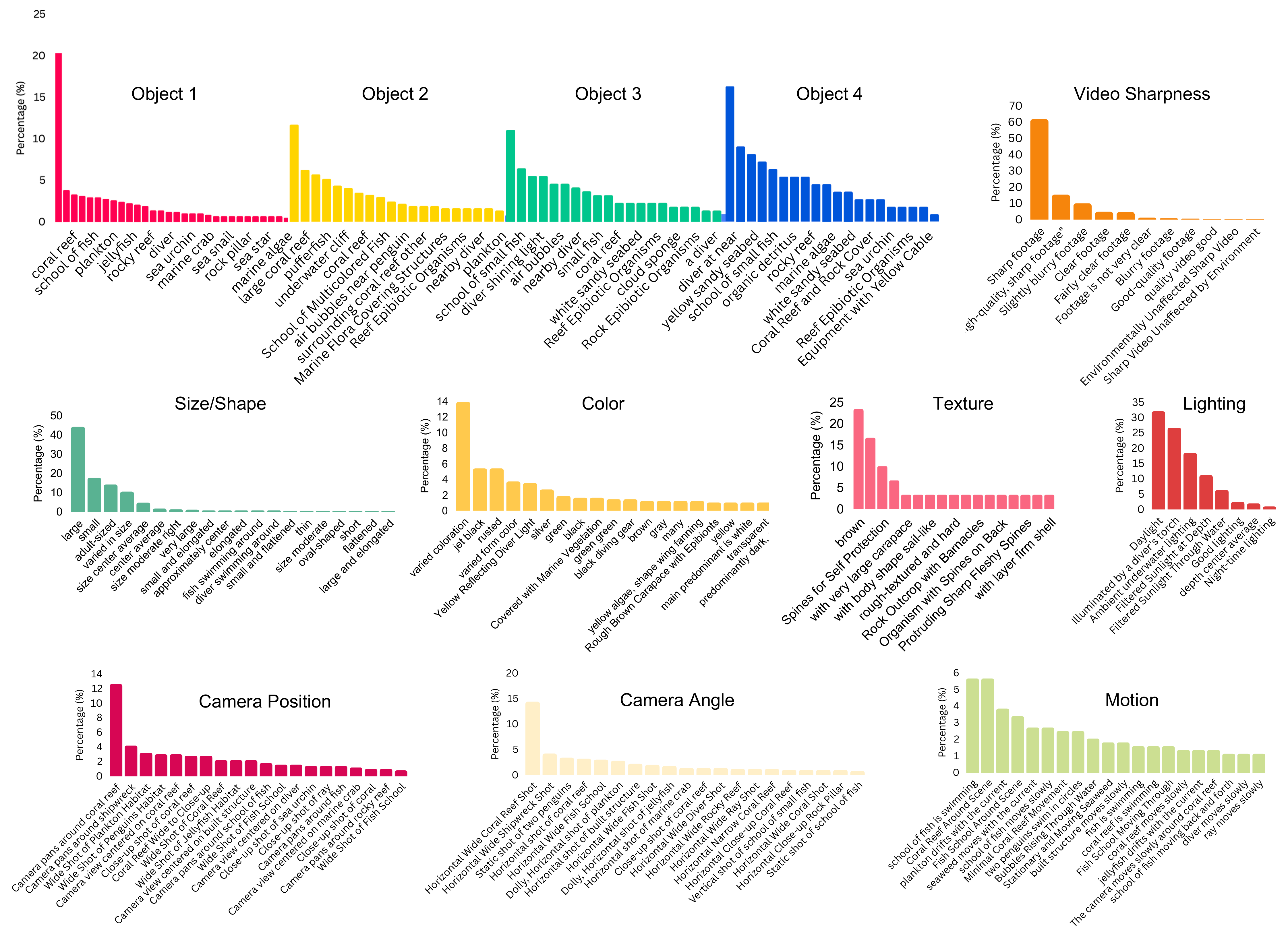}
   \caption{Statistics on the frequency of text-caption of \dataset. We show frequencies by instance object types (from object 1 to 4) \textbf{(Top)}, video quality \textbf{(Top Right)}, the size/shape, color, and texture of the central object \textbf{(Middle Left)}, camera motion (angle and position) \textbf{(Bottom Right)}, and object motion \textbf{(Bottom Left)}.}
   \label{fig:text_stats}
   \vspace{-0.2in}
\end{figure*}

\subsection{Statistics.}

We conduct a statistical analysis of the frequency of text captions of \dataset. We show frequencies by instance object types (from object 1 to 4), and fine-grained attributes in~\cref{fig:text_stats}. In our \dataset, there are 16,080 questions, including 12080 multiple choice questions and 4,000 open-ended questions for the collection of 2,000 videos.




\paragraph{Dataset Split.}

We split the dataset into a training set with 1300 videos, a validation set with 200 videos and a testing set with 500 videos.

\section{Evaluation Dimensions}
\begin{figure*}
    \centering
        \includegraphics[width=0.95\linewidth]{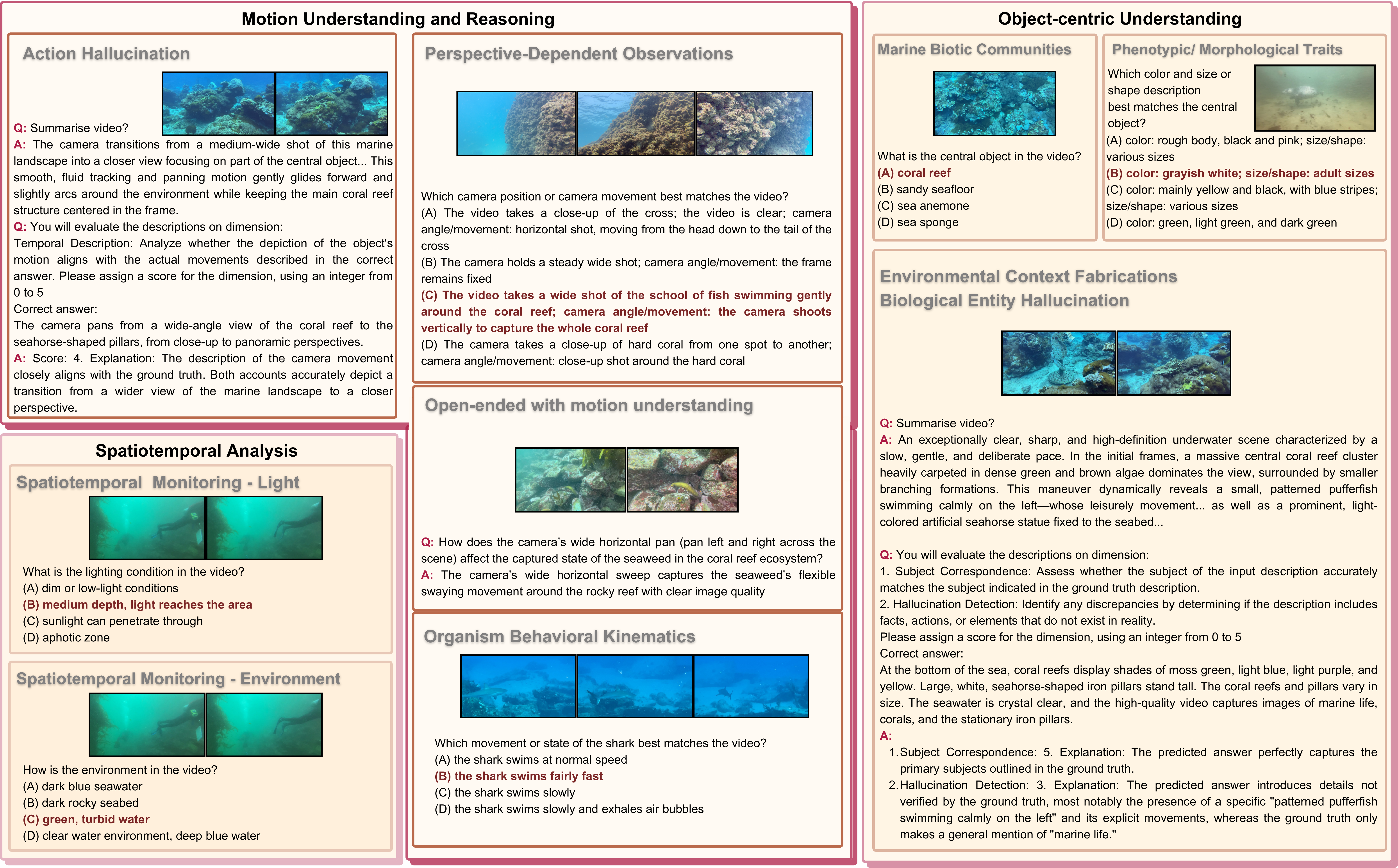}
        \caption{Overview of task dimensions of \dataset. See the supplementary material for the prompt templates.}
        \label{fig:dimensions}
\end{figure*}

To comprehensively evaluate model reasoning ability, we assess VLMs on multiple aspects, including how they summarize overall video content and how well they understand specific questions to test grounding and question-answering (QA) capabilities. We design two benchmarks focused on two key areas: multiple-choice marine-related QA and video captioning in~\cref{fig:dimensions}.

\subsection{Video-QA}
To comprehensively assess vision-language models in the marine domain, we systematically classify our evaluation protocol into five dimensions.

\begin{itemize}
    \item \textit{Marine Biotic Communities (MBC)} evaluates the model's capacity for multi-object detection and counting. Specifically, we aim to measure the precise identification of the central object as \textit{MBC-so} alongside up to three additional objects as \textit{MBC-mo} to reconstruct the marine biotic community.
    \item \textit{Phenotypic/Morphological Traits (PMT)} evaluates fine-grained feature extraction for taxonomic classification and biological health assessment. We target the accurate grounding and description of phenotypic traits like texture, size, shape, and color. 
    \item \textit{Organism Behavioral Kinematics (OBK)} assesses VLM’s capacity for spatiotemporal motion reasoning from marine video. We evaluate how effectively the model decodes complex motion/behavior patterns.
    \item \textit{Spatiotemporal Monitoring (SM)} measures VLM models’ understanding the environmental context. We assess the integration of environmental attributes, specifically under shifting lighting conditions as \textit{SM-l} and environments as \textit{SM-e}.
    \item \textit{Perspective-Dependent Observations (PDO)} evaluates the model's viewpoint invariance against dynamic camera pose. Specifically, we test the consistency of marine resource identification across diverse camera angles (horizontal, vertical, centered) and spatial positions (full scene, partial view).
    \item \textit{Open-Ended with Motion Understanding (OEMU)} assesses how VLMs captures camera viewpoint and objects. We target to object motion, central object and visible activity.
\end{itemize}

\subsection{Video Captioning with Motion Understanding}
Our captioning benchmark is used to rigorously measure the model's resistance to hallucination by assigning a dimension score from 0 to 5 to each generated prediction across the following three dimensions.

\begin{itemize}
    \item \textit{Biological Entity Hallucination (BEH)} evaluates whether the objects referenced in the generated description correctly match the objects specified in the ground truth.
    \item \textit{Action Hallucination (AH)} assesses whether the described object motion and camera pose are consistent with the actual movements.
    \item \textit{Environmental Context Fabrications (ECF)} identifies discrepancies by checking whether the generated description includes any facts or elements that are not present in reality.
\end{itemize}
\begin{figure}[t]
    \centering
    \includegraphics[width=0.9\linewidth]{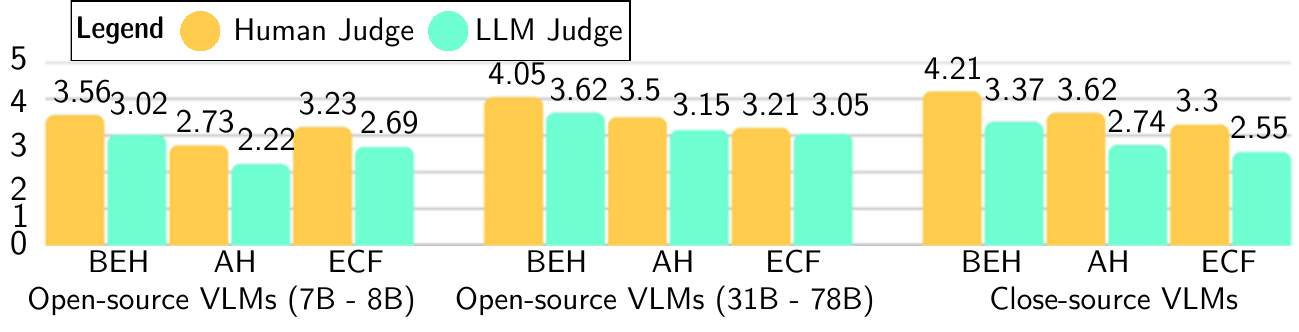}
    \caption{Human study. We recruit \textit{6 annotators} to score LLM predictions across \textit{100} video inputs, requiring \textit{180} total labor hours. We observe both human and LLM judges are consistent when evaluating open-ended QA, which confirms the effectiveness of using LLMs as judges for QA tasks.}
    \label{fig:humanstudy}
    \vspace{-0.2in}
\end{figure}
\section{Experiments}
Our goal is to evaluate how state-of-the-art video-LLMs handle VideoQA tasks, focusing on reasoning capabilities that account for motion dynamics, across both open-ended and multiple-choice questions. This section describes our experimental setup and model performance across evaluation dimensions in marine scenarios.

\subsection{Models}
We evaluate a broad range of proprietary and open-source VLMs for video QA and captioning. Our proprietary lineup includes frontier close-source models, i.e., Claude-4.8-Opus~\cite{claude48opus}, Grok-4.3~\cite{grok43}, Gemini-3.1-Pro~\cite{gemini31docs}, and Kimi K2.6~\cite{kimik26}. We further include open-source models, i.e., \textit{small VLMs} (InternVL3-8B~\cite{wang2025internvl3}, LLaVA-OneVision-7B~\cite{li2024llava}, Qwen2.5-VL-7B~\cite{wu2025qwen}, LLaVA-NeXT-Video~\cite{li2024llava1}, Video-LLaMA3~\cite{zhang2025videollama}) and \textit{larger VLMs} (Gemma4-31B~\cite{gemmateam2026gemma4}, Qwen2.5-VL-72B~\cite{wu2025qwen}, InternVL3-78B~\cite{wang2025internvl3}), using official Huggingface releases ranging from 7B to 78B parameters. We uniformly sample 8 frames per video as an ordered image sequence input. Empirically, we find 8 frames effectively capture camera motion while \textit{balancing} accuracy and token cost. All experiments for the open-source model are conducted using 6 NVIDIA GeForce RTX4090 D. We will release the prompts and hyperparameter configurations used for the public.

\subsection{Human Study}
To validate the LLM judge for open-ended QA, we conducted a human study with six domain experts, randomly selecting 100 questions across four open-ended QA types evaluating camera motion from visual attributes. Each question was rated by three experts, and scores were averaged. As shown in \cref{fig:humanstudy}, despite some gap between human and LLM (GPT-5.5) judges, human scoring is largely consistent with LLM scoring, confirming the effectiveness of using LLMs as judges for QA tasks. Small open-source models (7B-8B) struggle with AH.

\subsection{Multiple-choice Questions with Motion Understanding}
\begin{table*}[t]
\centering
\scalebox{0.7}{
\begin{tabular}{l r r r r >{\columncolor{red!15}} r r r r r}
\toprule
\textbf{Model} & \textbf{Backbone} & \textbf{MBC-so} & \textbf{MBC-mo} & \textbf{PMT}   & \textbf{PDO} & \textbf{OBK} & \textbf{SM-e} & \textbf{SM-l} & \textbf{OEMU} \\
\midrule
\multicolumn{9}{l}{\textbf{Open-source VLMs (7B - 8B)}} \\
\midrule
Video-Llama3-7B~\cite{zhang2025videollama} &  2D RoPE &  96.6 &  66.6 &  76.6 &  63.3 &  90 &  73.3 &  35.3 &  2.2\\
LLaVa-NeXT-Video~\cite{li2024llava1} &  CLIP & 80 &  56.7 &  40 &  60 &  83.3 &  23.3 &  11.8 &  2.14\\
LLaVA-OneVision-7B~\cite{li2024llava} & SigLIP/Qwen  & 90 &  63.3 &  60 &  53.3 &  80 &  83.3 &  29.4 &  2.11\\
Qwen2.5-VL-7B~\cite{wu2025qwen} & 2D RoPE/Qwen & 96.7 &  66.7 &  66.7 &  70 &  96.7 &  66.7 &  29.4 &  2.43\\
InternVL3-8B~\cite{wang2025internvl3} & V2PE/Qwen &  96.7 &  83.3 &  73.3 &  70 &  93.3 &  73.3 &  52.9 &  1.36\\
\midrule
Average & &   92.00 &  67.32 & 63.32 &  \textbf{63.32} & 88.66 & \textbf{63.98} & 31.76 & 2.04\\
\midrule
\multicolumn{9}{l}{\textbf{Open-source VLMs (31B - 78B)}} \\
\midrule
Gemma4-31B~\cite{gemmateam2026gemma4}                                     & SigLIP/Qwen                & 89.3 & 36.6 & 53.3 & 43.3 & 86.6 & 60   & 47.05 & 2.52 \\
Qwen2.5-VL-72B~\cite{wu2025qwen}                                 & ViT/Qwen2.5                & 90.5 & 76.6 & 80   & 46.7 & 93.3 & 66.7 & 66.7  & 2.47 \\
InternVL3-78B~\cite{wang2025internvl3}                                  & InternViT-6B/Qwen          & 91.2 & 63.3 & 70   & 46.6 & 83.4 & 53.3 & 29.4  & 1.98 \\
\midrule
Average & & 90.33 & 58.83 & 67.77 & 45.53 & 87.77 & 60 & 47.72 & 2.32 \\
\midrule
\multicolumn{9}{l}{\textbf{Close-source VLMs}} \\
\midrule
Claude-Opus-4.8~\cite{claude48opus} & - & 93.33 & 73.3 &  83.3 &  60 &  96.6 &  60 &  70.5 &  2.42\\
Grok-4.3~\cite{grok43} &  - & 93.33 & 70 &  76.6 &  53.3 &  93.3 &  60 &  70.5 &  2.58\\
Gemini-3.1-Pro~\cite{gemini31docs} &  - & 93.3 &  76.6 &  90 &  66.6 &  96.6 &  66.6 &  35.3 &  2.1\\
Kimi K2.6~\cite{kimik26} & - & 100 & 60 &  80 &  63.3 &  96.6 &  63.3 &  52.9 &  2.48\\
\midrule
Average & - &   \textbf{94.99} & \textbf{69.98} & \textbf{82.46} &  60.8 & \textbf{95.78} &  62.48 &  \textbf{57.3} &  \textbf{2.40}\\
\toprule
\end{tabular}}
\caption{Model performance on video question answering. This table reports model performance on our benchmark across MBC, PMT, OBK, SM and PDO metrics. Camera motion-specific metric is challenging for VLM, highlighted in \colorbox{red!15}{red}.}
\label{tab:qas}
\vspace{-0.2in}
\end{table*}

\begin{table}[t]
\centering
\scalebox{0.65}{
\begin{tabular}{lc>{\columncolor{red!15}}ccc>{\columncolor{red!15}}cc}
\toprule
\multirow{2}{*}{\textbf{Model}} & \multicolumn{3}{c}{\textsc{Static Camera}} & \multicolumn{3}{c}{\textsc{Moving Camera}} \\
\cmidrule(lr){2-4} \cmidrule(lr){5-7}
 & \textbf{BEH} & \textbf{AH} & \textbf{ECF} & \textbf{BEH} & \textbf{AH} & \textbf{ECF} \\
\midrule
\multicolumn{7}{l}{\textbf{Open-source VLMs (7B - 8B)}} \\
\midrule
Video-Llama3-7B~\cite{zhang2025videollama}     & 3.13 & 2.1  & 2.6  & 3.13 & 2.1  & 2.43 \\
LLaVa-NeXT-Video~\cite{li2024llava1}           & 2.9  & 2.71 & 2.4  & 2.43 & 1.72 & 2.42 \\
LLaVA-OneVision-7B~\cite{li2024llava}          & 3.1  & 2.29 & 3.14 & 3.1  & 1.85 & 3.14 \\
Qwen2.5-VL-7B~\cite{wu2025qwen}                & 3.43 & 2.71 & 2.71 & 3.14 & 2.58 & 2.57 \\
InternVL3-8B~\cite{wang2025internvl3}          & 2.57 & 1.86 & 2.3  & 2.71 & 1.86 & 2.42 \\
\midrule
Average & 3.03 & 2.33 & 2.63 & 2.9\small{\textcolor{WildStrawberry}{-0.13}}  & 2.02\small{\textcolor{WildStrawberry}{-0.31}} & 2.6\small{\textcolor{WildStrawberry}{-0.03}}  \\
\midrule
\multicolumn{7}{l}{\textbf{Open-source VLMs (31B - 78B)}} \\
\midrule
Gemma4-31B~\cite{gemmateam2026gemma4}          & 3.6  & 3.14 & 3.02 & 3.29 & 3.29 & 2.85 \\
Qwen2.5-VL-72B~\cite{wu2025qwen}               & 3.57 & 2.86 & 2.85 & 3.71 & 3.43 & 3.43 \\
InternVL3-78B~\cite{wang2025internvl3}         & 3.29 & 2.57 & 2.57 & 3.57 & 3.01 & 3.1  \\
\midrule
Average & \textbf{3.49} & \textbf{2.86} & \textbf{2.81} & \textbf{3.52}\small{\textcolor{WildStrawberry}{+0.03}} & \textbf{3.24}\small{\textcolor{WildStrawberry}{+0.38}} & \textbf{3.13}\small{\textcolor{WildStrawberry}{+0.32}} \\
\midrule
\multicolumn{7}{l}{\textbf{Close-source VLMs}} \\
\midrule
Claude-4.8-Opus~\cite{claude48opus}        & 3.14 & 3    & 2.3  & 3.29 & 2.86 & 2.43 \\
Grok-4.3~\cite{grok43}                     & 3.3  & 2.6  & 2.28 & 3.14 & 2.71 & 2.57 \\
Gemini-3.1-Pro~\cite{gemini31docs}             & 3.9  & 2.4  & 2.57 & 3.29 & 2.57 & 3.29 \\
Kimi K2.6~\cite{kimik26}                       & 3.6  & 3.11 & 2.43 & 3.14 & 3.1  & 2.28 \\
\midrule
Average & \textbf{3.49} & 2.78 & 2.4  & 3.22\small{\textcolor{WildStrawberry}{-0.27}} & 2.81\small{\textcolor{WildStrawberry}{+0.03}} & 2.64\small{\textcolor{WildStrawberry}{+0.24}} \\
\bottomrule
\end{tabular}}
\caption{Comparison of models across static and moving camera settings. Camera motion-specific metric is challenging for VLM, highlighted in \colorbox{red!15}{red}.}
\label{tab:static_moving_camera}
\vspace{-0.2in}
\end{table}

\paragraph{Observation \#1: Proprietary models outperform open-source models across most downstream benchmarks.}
We find that closed-source models outperform open-source models across most evaluation dimensions like MBC-so, MBC-mo, PMT, OBK, SM-l and OEMU in~\cref{tab:qas} because closed-source models often train on larger and private datasets rather than open-source datasets.

\paragraph{Observation \#2: While all VLM models exhibit robust capabilities in object motion understanding, they often struggle to capture complex camera motion.}
The object motion scores OBK of all VLMs are high (average open-source model: 88.66\% (7B-8B VLMs), 87.77\% (31B-78B VLMs), average close-source model: 95.78\%) in~\cref{tab:qas}. However, the PDO scores are dramatically low (average 7B-8B and 31B-78B open-source models: 63.32\% and 45.53\%, average close-source model: 60.8\%), respectively, and for the close-source group, it is the second lowest across all metrics. This explains why object tracking performs fairly well in which object recognition achieves strong performance with 92\% and 90.33\% for open-source models 7B-8B and 31B-78B, and 94.99\% for proprietary models in MBC-so. But when the camera changes, the visual encoder struggles to extract discriminative features, resulting in poor camera motion reasoning in autoregressive language modeling.

\paragraph{Observation \#3: The reasoning ability of open-source models 7B-8B is strong, but limited in camera-motion perception.}
Although the close-source group dominates across the overall average, in the camera motion PDO metric, open-source models like Qwen2.5-VL-7B (70\%) and InternVL3-8B (70.0\%) outperforms close-source models like Grok-4.3 (53.3\%) and Claude-Opus (60.0\%) in~\cref{tab:qas}. This suggests that Qwen2.5-VL leverages an upgraded Dynamic Resolution ViT, while InternVL3 adopts the Dynamic High-Resolution (DHR) image/frame decomposition strategy. Both enable dynamic token pruning, which results in more effective temporal representation than vanilla average temporal pooling. On the contrary, 31B–78B VLMs often produce detailed captions, but they struggle to predict complex camera motion.

\begin{figure}[t]
    \centering
    \includegraphics[width=1\linewidth]{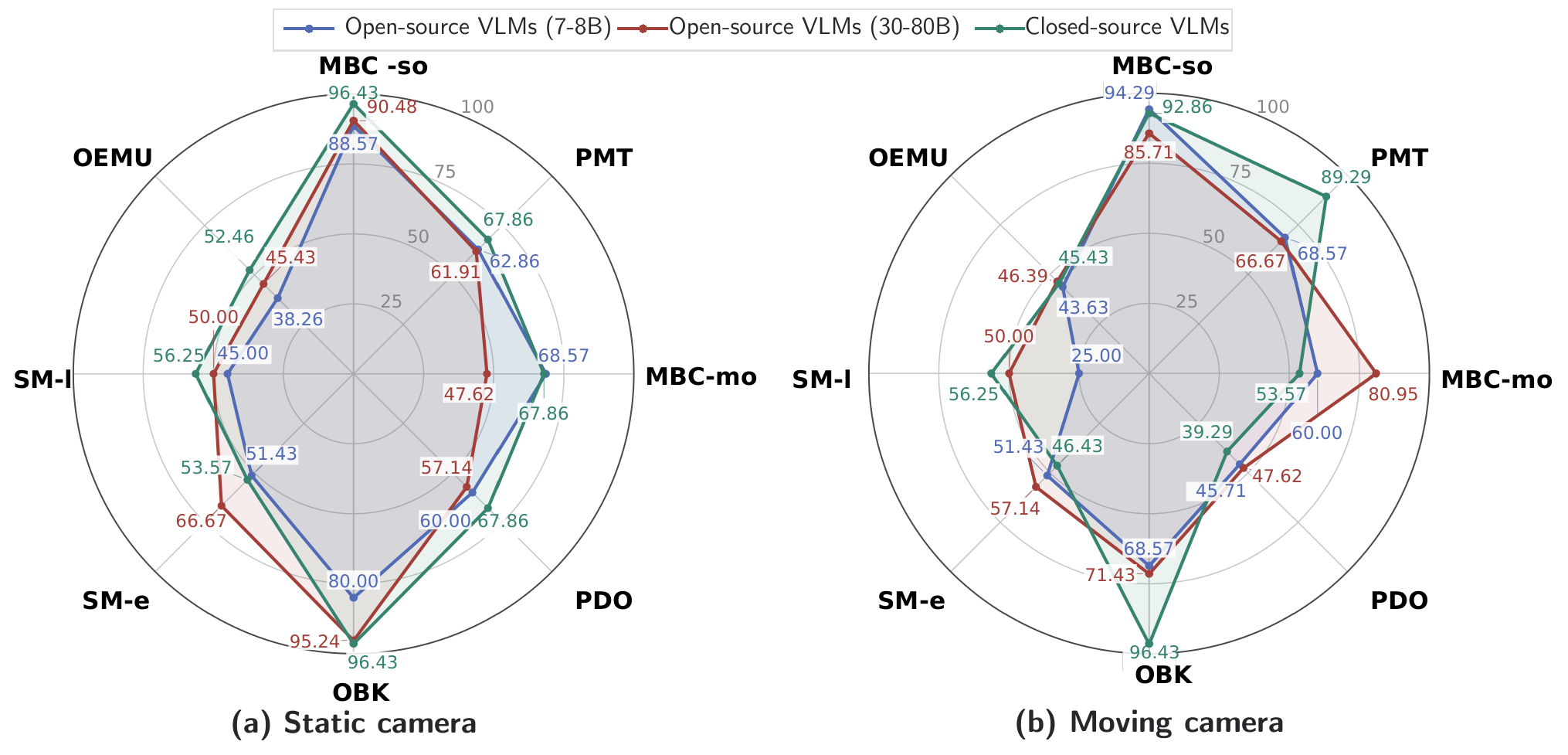}
    \caption{Model performance of VLMs on static and moving camera settings. }
    \label{fig:camera}
    \vspace{-0.2in}
\end{figure}

\subsection{Open-ended Questions}
\paragraph{Observation \#4: For video summarization, VLMs struggle with camera motion, failing to reason in the static-camera setting.}
~\cref{tab:static_moving_camera} shows that, for moving cameras, both open-source and closed-source models achieve superior performance compared to static cameras in open-ended QA. However, VLMs 7B-8B struggle with reasoning capability (ranging from 2.02 to 2.9) across all metrics under camera movement. Thus, we believe VLMs still struggle with camera motion (even static camera), leading to require additional training data with rich camera-motion attributes for finetuning. Taking a closer look at multiple-choice questions, ~\cref{fig:camera} shows VLMs struggle on SM-l, SE-e, and OEMU when the camera is moving, about 50\%. In contrast, MBC-so achieves good performance about 90\% in both settings. For OBK, performance is slightly higher for moving cameras than for static cameras, yet SM-l remains challenging for VLMs on moving-camera inputs.
\section{Conclusion}
In this work, we introduced a sequential multi-trajectory expert annotation and refinement strategy, using a human-in-the-loop multi-agent data engine for curating high-quality, motion-centric video-text data for marine environments. Using this strategy, we built a benchmark of 2,000 marine videos with 16,080 fine-grained question-answer pairs, annotated across 13 attributes covering key marine video challenges. We further constructed a comprehensive evaluation suite spanning nine dimensions across open-ended and multiple-choice QA. Our results show both open-source and closed-source models struggle most with camera motion dynamics. We hope this benchmark advances underwater video understanding.

\section*{Acknowledgements.}
This project is funded by the Environment and Conservation Fund (Project No.: ECWW26IS07 235/2025) of the Hong Kong Special Administrative Region Government. We thank all the volunteer students who helped do the annotations and validation.

{
    \small
    \bibliographystyle{ieeenat_fullname}
    \bibliography{main}
}

\appendix

\section{Introduction}
In this supplementary material, we present the copyright of our dataset in \textit{‘Copyright’}. The prompts for the caption engine are illustrated in \textit{‘Data Engine’}. We also present a prompt for QA generation in \textit{‘Multiple-choice Generation’}. Evaluation dimensions for open-ended QA with motion understanding, aimed at detecting hallucinations in predicted answers, are introduced in \textit{‘Evaluation Dimensions of Open-ended’}.

\section{Copyright}
\label{sec:copyright}
This dataset is released for research purposes only. The dataset uses or is derived from data licensed from MVK~\cite{truong2023marine} and MSC~\cite{truong2025msc} datasets. Prior to releasing this dataset, we contacted the respective data owners of MVK and MSC and obtained their permission to use their datasets for research use. In this submission, we release a small subset of the dataset, consisting of 500 videos. A full-scale version of the dataset will be publicly available after acceptance.

\section{Dataset Details – Collection \& Generation}
In this section, we provide additional details on dataset construction, including the captioning engine, QA generation, and evaluation metrics.

\begin{table}[t]
\centering
\begin{tabular}{p{\linewidth}}
\toprule
\textbf{Instruction} \\ 
\midrule
You are given 8 frames sampled chronologically from a single underwater video.

Treat all 8 frames as one video sequence, not as separate images.

Write one complete English video-captioning sentence that contains fewer than 80 words, describes the central object's identity, texture or pattern, color, size, and shape, includes up to three secondary objects, details the environment and lighting, mentions the video quality, describes the object motion, and defines the camera position and camera angle separately.\\
\bottomrule
\end{tabular}
\caption{Prompt for Captioner in caption generation, belonging to step 2 of the caption engine.}
\label{tab:captioner}
\end{table}

\subsection{Data Engine}
\label{sec:data_engine}
We introduce a three-stage, human-in-the-loop caption engine that utilizes a multi-agent framework to generate high-quality video captions. First, domain experts will annotate attributes in correspondence with marine video. Second, we use a sequence of video frames as input to an LLM to generate captions. We use the prompt in~\cref{tab:captioner} for the captioning agent. Then, we use a prompt in~\cref{tab:reviewer} with another LLM as a reviewer to score the semantic alignment between the synthetic caption and the manually annotated attributes. Third, an LLM refiner generates the final caption using the prompt in~\cref{tab:refiner}.

\subsection{Multiple-choice Generation}
\label{sec:mc}
For multiple-choice questions, we use fine-grained attributes as answer candidates. We then use an LLM prompt in~\cref{tab:QA1} with rules in~\cref{tab:QA2} to generate the final answer choices.

\section{Evaluation Dimensions of Open-ended QA}
\label{sec:dim}
\paragraph{Open-ended with motion understanding.} We use video captions as facts to create question-answer pairs, evaluating the VLM’s ability to understand camera motions. We use a LLM judge with promtps in~\cref{tab:open-endedQA} to evaluate a model’s capability, aiming to decode the functional relationship between camera movement and the corresponding object's attributes.

\paragraph{Video captioning.}
We aim to measure the model’s resistance to hallucinations in~\cref{tab:captioning}. Specifically, we evaluate whether the objects referenced in the generated descriptions correctly match the objects specified in the ground truth. For action hallucination, we assess whether the described object motion in the predicted answer and reference answer is consistent with the actual movements, using this prompt. Finally, we also use the mentioning prompt to identify discrepancies by checking whether the generated description includes any facts or elements that are not present in reality.

\begin{table*}[htbp]
\centering
\begin{tabular}{p{\linewidth}}
\toprule
\textbf{Instructions} \\
\midrule
You are an advanced intelligent chatbot specialized in evaluating the factual accuracy of generative outputs related to video descriptions. Your primary task is to meticulously compare a ``Predicted Answer'' with the provided ``Ground Truth'' (human-annotated attributes) and assess their factual consistency. \\
\\
You will evaluate the alignment between the Predicted Answer and the Ground Truth across specific visual and contextual dimensions. \\
\\
\textbf{\large{EVALUATION DIMENSIONS}} \\
Evaluate alignment across these 8 dimensions:
\begin{enumerate}[label=\arabic*., itemsep=0pt, topsep=3pt]
    \item \textbf{Objects} (Presence \& count of up to 4 main objects)
    \item \textbf{Object Attributes} (Texture, size, shape, color of central objects)
    \item \textbf{Environment} (Setting/location)
    \item \textbf{Lighting} (Type/quality of illumination)
    \item \textbf{Video Clarity} (Resolution/focus)
    \item \textbf{Motion/Behavior} (Actions/movement)
    \item \textbf{Camera Angle} (Perspective, e.g., low-angle, eye-level)
    \item \textbf{Camera Position} (Framing/movement, e.g., close-up, panning)
\end{enumerate} \\
\textbf{\large{SCORING RUBRIC (0-5 Integer Scale)}} \\
Please assign an average score for all these dimensions, using an integer from 0 to 5, where 5 indicates a perfect match and 0 signifies no relevance at all. \\
\\
\textbf{\large{INPUT DATA}} \\
\begin{itemize}[label=$\bullet$, itemsep=0pt, topsep=3pt]
    \item \textbf{Ground Truth (Human Annotation):} [Insert Human Annotation/Attributes Here]
    \item \textbf{Predicted Answer:} [Insert Candidate Output Here]
\end{itemize} \\
\bottomrule
\end{tabular}
\caption{Prompt for reviewer in the caption engine.}
\label{tab:reviewer}
\end{table*}

\begin{table*}[htbp]
\centering
\begin{tabular}{p{\linewidth}}
\toprule
\textbf{Instructions} \\
\midrule
You are an expert video annotation editor. Your task is to correct a flawed ``Predicted Caption'' by incorporating the missing or incorrect details found in the provided ``Ground Truth Attributes''. \\
\\
\textbf{\large{INPUT DATA}}
\begin{itemize}[label=$\bullet$, itemsep=0pt, topsep=3pt]
    \item \textbf{Original Predicted Caption:} [Insert original caption here]
    \item \textbf{Ground Truth Attributes:} [Insert human annotations/attributes here]
    \item \textbf{Lower-Scoring Caption:} [List caption that their scores less than 3.5]
\end{itemize} \\\\

\textbf{\large{INSTRUCTIONS}}
\begin{enumerate}[label=\arabic*., itemsep=0pt, topsep=3pt]
    \item Review the errors and omissions in the Predicted Caption based on the Ground Truth.
    \item Rewrite the caption to seamlessly integrate the correct attributes (objects, textures, sizes, shapes, colors, environment, lighting, motion, and camera details).
    \item Do not hallucinate. Ensure every detail in the corrected caption matches the Ground Truth perfectly.
    \item Keep the caption concise, natural-sounding, and visually descriptive.
\end{enumerate} \\
\textbf{\large{OUTPUT FORMAT}} \\
\textbf{\large{Corrected Caption:}} $[$[Insert your newly edited, factually flawless caption here$]$ \\
\bottomrule
\end{tabular}
\caption{Prompt for refiner in the caption engine.}
\label{tab:refiner}
\end{table*}

\begin{table*}[htbp]
\centering
\begin{tabular}{p{\linewidth}}
\toprule
\textbf{Instructions} \\
\midrule
You are an expert dataset creator for Video Question-Answering in the marine biology domain. Your task is to transform a row of raw video annotation data (Ground Truth) into a dynamic list of high-quality QA pairs following a strict JSONL schema. \\
\\
\textbf{\large{INPUT}} \\
You will receive a JSON object representing the Ground Truth (GT) of a single video. It contains attributes such as: \texttt{video\_id}, \texttt{collection}, \texttt{central\_object}, \texttt{color}, \texttt{size\_shape}, \texttt{objects\_present}, \texttt{camera\_position}, \texttt{camera\_angle}, \texttt{movement\_behavior}, \texttt{environment}, and \texttt{lighting}. \\
\\
\textbf{\large{OUTPUT FORMAT}} \\
You must output the results in \textbf{JSON Lines (JSONL) format}. This means you will output between 6 to 9 individual JSON objects (depending on the presence of optional attributes). \textbf{Each object must be on its own single line}, with no trailing commas between lines, and they MUST NOT be wrapped in a JSON array. \\
\\
Each JSON object MUST strictly follow this schema: \\
\\
\texttt{\{} \\
\texttt{~~"question\_id": "\{collection\}\_\_\{video\_file\_name\_without\_ext\}\_\_q\{index\}",} \\
\texttt{~~"id": "\{collection\}\_\_\{video\_file\_name\_without\_ext\}",} \\
\texttt{~~"task": "qa",} \\
\texttt{~~"type": "<type\_of\_question>",} \\
\texttt{~~"question\_format": "multiple\_choice" OR "open\_ended",} \\
\texttt{~~"collection": "\{collection\}",} \\
\texttt{~~"video\_id": "\{video\_id\}",} \\
\texttt{~~"video\_file": "\{video\_file\_name\_with\_ext\}",} \\
\texttt{~~"video\_path": "\{collection\}/\{video\_file\_name\_with\_ext\}",} \\
\texttt{~~"central\_object": "\{central\_object\}",} \\
\texttt{~~"question": "<The standalone question>",} \\
\texttt{~~"prompt": "<For MCQ: The question + options (A) (B) (C) (D) +} \\
\texttt{~~~~~~~~~~~~'Answer with the option letter only.' | For open-ended: Just the question>",} \\
\texttt{~~"options": [} \\
\texttt{~~~~\{"label": "A", "text": "<Option A text>\"\},} \\
\texttt{~~~~\{"label": "B", "text": "<Option B text>\"\},} \\
\texttt{~~~~\{"label": "C", "text": "<Option C text>\"\},} \\
\texttt{~~~~\{"label": "D", "text": "<Option D text>\"\}} \\
\texttt{~~] (Empty array [] for open-ended),} \\
\texttt{~~"answer": "<A, B, C, or D> (Empty string for open-ended)",} \\
\texttt{~~"answer\_text": "<Text of the correct answer or reference answer>",} \\
\texttt{~~"evidence\_attributes": \{ "<key>": "<gt\_value\_used>" \}} \\
\texttt{\}} \\
\bottomrule
\end{tabular}
\caption{Prompt for QA Generation. We define the input and output contents.}
\label{tab:QA1}
\end{table*}

\begin{table*}[htbp]
\centering
\begin{tabular}{p\linewidth}
\toprule
\textbf{Instructions} \\
\midrule
\textbf{\large{QUESTION GENERATION RULES}} \newline\newline
\textbf{A. Mandatory Questions (Always generate these 6):} \newline
1. type: "central\_object" (MCQ) \newline
\hspace*{3mm} - Question: "What is the central object in the video?" \newline
\hspace*{3mm} - Correct Answer: The exact value of \texttt{central\_object}. \newline
2. type: "central\_object\_appearance" (MCQ) \newline
\hspace*{3mm} - Question: "Which color and size or shape description best matches the [insert central\_object here]?" \newline
\hspace*{3mm} - Correct Answer Format: "color: [GT color]; size/shape: [GT size\_shape]". \newline
3. type: "object\_detect" (MCQ) \newline
\hspace*{3mm} - Question: "Which option lists only objects present in the video? An option is incorrect if it includes any object not shown." \newline
\hspace*{3mm} - Correct Answer: A comma-separated list combining the \texttt{central\_object} AND all secondary objects (e.g., \texttt{object\_2}, \texttt{object\_3}). If there are NO secondary objects in the GT, the correct answer is simply the \texttt{central\_object} by itself. Do NOT use any fallback text. \newline
4. type: "camera\_motion" (MCQ) \newline
\hspace*{3mm} - Question: "Which camera position or camera movement best matches the video?" \newline
\hspace*{3mm} - Correct Answer Format: "camera position/view: [GT camera\_position]; camera angle/movement: [GT camera\_angle]". \newline
5. type: "camera\_object\_open\_ended" (open\_ended) \newline
\hspace*{3mm} - Question: "How do the camera movement and viewpoint capture the [insert central\_object here] and the visible object activity in the video?" \newline
\hspace*{3mm} - Reference Answer: Construct a fluent sentence explicitly combining \texttt{central\_object}, \texttt{movement\_behavior}, \texttt{camera\_position}, and \texttt{camera\_angle}. \newline
6. type: "caption" (open\_ended) \newline
\hspace*{3mm} - Question: "Write one complete English video-captioning sentence that contains fewer than 80 words, describes the central object's identity, texture or pattern, color, size, and shape, includes up to three secondary objects, details the environment and lighting, mentions the video quality, describes the object motion, and defines the camera position and camera angle separately." \newline
\hspace*{3mm} - Reference Answer: Use video captions from the dataset GT \newline\newline
\textbf{B. Conditional Questions (ONLY generate if the data exists in GT):} \newline
7. type: "central\_object\_motion" (MCQ) \newline
\hspace*{3mm} - CONDITION: Only generate this question if \texttt{movement\_behavior} describes the movement or state of the \texttt{central\_object}. \newline
\hspace*{3mm} - Question: "Which movement or state of the [central\_object], best matches the video?" \newline
\hspace*{3mm} - Correct Answer: The movement/state of the \texttt{central\_object} only. \newline
\hspace*{3mm} - IMPORTANT: All answer options must describe the same \texttt{central\_object}. Do not use other objects or species as the subject of any option. \newline
8. type: "environment" (MCQ) \newline
\hspace*{3mm} - CONDITION: ONLY generate this if \texttt{environment} is NOT empty or null. \newline
\hspace*{3mm} - Question: "How is the environment in the video?" \newline
\hspace*{3mm} - Correct Answer: The exact value of \texttt{environment}. \newline
9. type: "lighting" (MCQ) \newline
\hspace*{3mm} - CONDITION: ONLY generate this if \texttt{lighting} is NOT empty or null. \newline
\hspace*{3mm} - Question: "How is the lighting in the video?" \newline
\hspace*{3mm} - Correct Answer: The exact value of \texttt{lighting}. \newline

Here is the Ground Truth data for the current video. Please evaluate the conditional rules, generate the appropriate number of JSON objects with correct sequential indexing (\texttt{q1}, \texttt{q2}, etc. without skipping numbers), and compile them directly into a valid \textbf{JSONL format} exactly like a \texttt{.jsonl} file (one valid JSON object per line). \\
\bottomrule
\end{tabular}
\caption{Prompt for QA Generation. We define rules to generate multiple-choices.}
\label{tab:QA2}
\end{table*}

\begin{table*}[htbp]
\centering\begin{tabular}{p{\linewidth}}
\toprule
\textbf{Instructions} \\
\midrule
You are evaluating an open-ended Video Question Answering response. \\
\\
Use only the structured ground-truth evidence supplied below. The reference answer is an example of a correct response, not a string that must be matched word-for-word. Accept accurate paraphrases and different sentence structures. Do not reward unsupported detail, verbosity, or stylistic similarity. \\
\\
Score each criterion with an integer from 0 to 5:
\begin{enumerate}[label=\arabic*., itemsep=0pt, topsep=3pt]
    \item \textbf{Camera Motion and Viewpoint (weight 35\%):} Does it correctly describe the camera position, framing, angle, and movement contained in the GT?
    \item \textbf{Object Activity (weight 30\%):} Does it correctly describe the annotated movement, behavior, or stationary state of the central/supporting objects?
    \item \textbf{Camera-Object Relationship (weight 35\%):} Does it explicitly explain how the camera movement/viewpoint records, follows, reveals, or frames the object activity?
\end{enumerate} \\
Scoring anchors:
\begin{itemize}[label=--, itemsep=0pt, topsep=3pt]
    \item \textbf{5:} fully correct and complete for that criterion.
    \item \textbf{4:} correct main content with one minor omission.
    \item \textbf{3:} partially correct but misses an important detail.
    \item \textbf{2:} limited relevant content or includes a meaningful error.
    \item \textbf{1:} minimally related.
    \item \textbf{0:} absent, completely incorrect, or non-responsive.
\end{itemize} \\
Return only JSON matching the requested schema. \\
\\
\textbf{\large{Question}} \\
\{question\} \\
\\
\textbf{\large{Reference answer}} \\
\{reference\_answer\} \\
\\
\textbf{\large{Predicted answer}} \\
\{predicted\_answer\} \\
\bottomrule
\end{tabular}
\caption{Prompt for Open-ended video QA evaluation, i.e., open-ended with motion understanding (QEMU).}
\label{tab:open-endedQA}
\end{table*}

\begin{table*}[htbp]
\centering
\begin{tabular}{p{\linewidth}}
\toprule
\textbf{Instructions} \\
\midrule
You are an advanced intelligent chatbot specialized in evaluating the factual accuracy of generative outputs related to object-based video descriptions. \\
\\
Your primary task is to meticulously compare the predicted answer with the correct answer and assess their factual consistency. To accomplish this, evaluate the descriptions across three key dimensions:
\begin{enumerate}[label=\arabic*., itemsep=0pt, topsep=3pt]
    \item \textbf{Subject Correspondence:} Assess whether the subject of the input description accurately matches the subject indicated in the ground truth description.
    \item \textbf{Temporal Description:} Analyze whether the depiction of the object's motion aligns with the actual movements described in the correct answer.
    \item \textbf{Hallucination Detection:} Identify any discrepancies by determining if the description includes facts, actions, or elements that do not exist in reality, such as imaginative interpretations or incorrect inferences.
\end{enumerate} \\
Assign an integer score from 0 to 5 for each dimension, where 5 indicates a perfect match and 0 signifies no relevance at all. For Hallucination Detection, 5 means no hallucination and 0 means severe hallucination. Accompany each score with a brief explanation. \\
\\
Return only JSON matching the requested schema. \\
\\
\textbf{\large{Correct answer:}} \\
\{correct\_answer\} \\
\\
\textbf{\large{Predicted answer:}} \\
\{predicted\_answer\} \\
\bottomrule
\end{tabular}
\caption{Prompt for video caption evaluation, including biological entity hallucination, action hallucination and environmental context fabrications.}
\label{tab:captioning}
\end{table*}

\end{document}